\documentclass[10pt,twocolumn,letterpaper]{article}

\usepackage{iccv}
\makeatletter
\@namedef{ver@everyshi.sty}{}
\makeatother

\usepackage[dvipsnames,svgnames,x11names]{xcolor}
\usepackage{tikz}
\usetikzlibrary{arrows.meta,shapes,calc,matrix,fit,positioning,backgrounds,decorations.markings}
\usepackage{pgfplots}
\usepackage{pgfplotstable}
\pgfplotsset{compat=1.9}
\usepackage{xstring}

\usepgfplotslibrary{external}
\IfBeginWith*{\jobname}{fig/extern/}{\finalcopy}{}

\tikzstyle{every picture}+=[
	remember picture,
	every text node part/.style={align=center},
	every matrix/.append style={ampersand replacement=\&},
]
\tikzstyle{tight} = [inner sep=0pt,outer sep=0pt]
\tikzstyle{node}  = [draw,circle,tight,minimum size=12pt,anchor=center]
\tikzstyle{op}    = [draw,circle,tight]
\tikzstyle{dot}   = [fill,draw,circle,inner sep=1pt,outer sep=0]
\tikzstyle{pt}    = [fill,draw,circle,inner sep=1.5pt,outer sep=.2pt]
\tikzstyle{box}   = [draw,thick,rectangle,inner sep=3pt]
\tikzstyle{high}  = [black!60]
\tikzstyle{group} = [high,box,opacity=.5]
\tikzstyle{rectc} = [tight,transform shape]
\tikzstyle{rect}  = [rectc,anchor=south west]

\newcommand{\leg}[1]{\addlegendentry{#1}}

\tikzset{every mark/.append style={solid}}
\pgfplotsset{%smooth,
	grid=both, width=\columnwidth, try min ticks=5,
	every axis/.append style={font=\small},
	every axis plot/.append style={thick,mark=none,mark size=1,tension=0.18},
	legend cell align=left, legend style={fill opacity=0.8},
	xticklabel={\pgfmathprintnumber[assume math mode=true]{\tick}},
	yticklabel={\pgfmathprintnumber[assume math mode=true]{\tick}},
	nodes near coords math/.style={
		nodes near coords={\pgfmathprintnumber[assume math mode=true]{\pgfplotspointmeta}},
	},
}

\pgfplotsset{
	dash/.style={mark=o,dashed,opacity=0.6},
	dott/.style={mark=o,dotted,opacity=0.6},
	nolim/.style={enlargelimits=false},
	plain/.style={every axis plot/.append style={},nolim,grid=none},
}

\makeatletter
\renewcommand\paragraph{\@startsection{paragraph}{4}{\z@}{1ex}{-1em}{\normalfont\normalsize\bfseries}}
\makeatother
\usepackage{times}
\usepackage{epsfig}
\usepackage{graphicx}
\usepackage{amsmath}
\usepackage{amssymb}
\usepackage{multirow}
\usepackage{subcaption}
\usepackage{enumitem}
\usepackage{booktabs}
\usepackage{balance}
\usepackage[accsupp]{axessibility}
\usepackage{pifont}% http://ctan.org/pkg/pifont
\usepackage[pagebackref=true,breaklinks=true,letterpaper=true,colorlinks,bookmarks=false]{hyperref}

\iccvfinalcopy % *** Uncomment this line for the final submission

\def\iccvPaperID{xx} % *** Enter the ICCV Paper ID here

\ificcvfinal\fi

\begin{document}

%%%%%%%%% TITLE
\title{Zero-Shot and Few-Shot Video Question Answering with Multi-Modal Prompts}

\author{Deniz Engin$^{1}$ \qquad  Yannis Avrithis$^{2}$ \\ \\
$^{1}$Inria, Univ Rennes, CNRS, IRISA \\ $^{2}$Institute of Advanced Research in Artificial Intelligence (IARAI)}

\maketitle
\ificcvfinal\thispagestyle{empty}\fi

% !TEX root = ../paper.tex

\newcommand{\head}[1]{{\smallskip\noindent\textbf{#1}}}
\newcommand{\alert}[1]{{\color{red}{#1}}}
\newcommand{\sm}{\scriptsize}
\newcommand{\eq}[1]{(\ref{eq:#1})}

\newcommand{\Th}[1]{\textsc{#1}}
\newcommand{\mr}[2]{\multirow{#1}{*}{#2}}
\newcommand{\mc}[2]{\multicolumn{#1}{c}{#2}}
\newcommand{\tb}[1]{\textbf{#1}}
\newcommand{\ch}{\checkmark}

\newcommand{\red}[1]{{\color{red}{#1}}}
\newcommand{\blue}[1]{{\color{blue}{#1}}}
\newcommand{\green}[1]{{\color{green}{#1}}}
\newcommand{\gray}[1]{{\color{gray}{#1}}}

\newcommand{\citeme}[1]{\red{[XX]}}
\newcommand{\refme}[1]{\red{(XX)}}

\newcommand{\fig}[2][1]{\includegraphics[width=#1\linewidth]{fig/#2}}
\newcommand{\figh}[2][1]{\includegraphics[height=#1\linewidth]{fig/#2}}

%--------------------------------------------------------------------

\newcommand{\tran}{^\top}
\newcommand{\mtran}{^{-\top}}
\newcommand{\zcol}{\mathbf{0}}
\newcommand{\zrow}{\zcol\tran}

\newcommand{\ind}{\mathbbm{1}}
\newcommand{\expect}{\mathbb{E}}
\newcommand{\nat}{\mathbb{N}}
\newcommand{\zahl}{\mathbb{Z}}
\newcommand{\real}{\mathbb{R}}
\newcommand{\proj}{\mathbb{P}}
\newcommand{\prob}{\mathbf{Pr}}
\newcommand{\normal}{\mathcal{N}}

\newcommand{\mif}{\textrm{if}\ }
\newcommand{\other}{\textrm{otherwise}}
\newcommand{\minimize}{\textrm{minimize}\ }
\newcommand{\maximize}{\textrm{maximize}\ }
\newcommand{\st}{\textrm{subject\ to}\ }

\newcommand{\id}{\operatorname{id}}
\newcommand{\const}{\operatorname{const}}
\newcommand{\sgn}{\operatorname{sgn}}
\newcommand{\var}{\operatorname{Var}}
\newcommand{\mean}{\operatorname{mean}}
\newcommand{\trace}{\operatorname{tr}}
\newcommand{\diag}{\operatorname{diag}}
\newcommand{\vect}{\operatorname{vec}}
\newcommand{\cov}{\operatorname{cov}}
\newcommand{\sign}{\operatorname{sign}}
\newcommand{\prj}{\operatorname{proj}}

\newcommand{\softmax}{\operatorname{softmax}}
\newcommand{\clip}{\operatorname{clip}}

\newcommand{\defn}{\mathrel{:=}}
\newcommand{\peq}{\mathrel{+\!=}}
\newcommand{\meq}{\mathrel{-\!=}}

\newcommand{\floor}[1]{\left\lfloor{#1}\right\rfloor}
\newcommand{\ceil}[1]{\left\lceil{#1}\right\rceil}
\newcommand{\inner}[1]{\left\langle{#1}\right\rangle}
\newcommand{\norm}[1]{\left\|{#1}\right\|}
\newcommand{\abs}[1]{\left|{#1}\right|}
\newcommand{\frob}[1]{\norm{#1}_F}
\newcommand{\card}[1]{\left|{#1}\right|\xspace}
\newcommand{\diff}{\mathrm{d}}
\newcommand{\der}[3][]{\frac{d^{#1}#2}{d#3^{#1}}}
\newcommand{\pder}[3][]{\frac{\partial^{#1}{#2}}{\partial{#3^{#1}}}}
\newcommand{\ipder}[3][]{\partial^{#1}{#2}/\partial{#3^{#1}}}
\newcommand{\dder}[3]{\frac{\partial^2{#1}}{\partial{#2}\partial{#3}}}

\newcommand{\wb}[1]{\overline{#1}}
\newcommand{\wt}[1]{\widetilde{#1}}

\def\xssp{\hspace{1pt}}
\def\ssp{\hspace{3pt}}
\def\msp{\hspace{5pt}}
\def\lsp{\hspace{12pt}}

\newcommand{\cA}{\mathcal{A}}
\newcommand{\cB}{\mathcal{B}}
\newcommand{\cC}{\mathcal{C}}
\newcommand{\cD}{\mathcal{D}}
\newcommand{\cE}{\mathcal{E}}
\newcommand{\cF}{\mathcal{F}}
\newcommand{\cG}{\mathcal{G}}
\newcommand{\cH}{\mathcal{H}}
\newcommand{\cI}{\mathcal{I}}
\newcommand{\cJ}{\mathcal{J}}
\newcommand{\cK}{\mathcal{K}}
\newcommand{\cL}{\mathcal{L}}
\newcommand{\cM}{\mathcal{M}}
\newcommand{\cN}{\mathcal{N}}
\newcommand{\cO}{\mathcal{O}}
\newcommand{\cP}{\mathcal{P}}
\newcommand{\cQ}{\mathcal{Q}}
\newcommand{\cR}{\mathcal{R}}
\newcommand{\cS}{\mathcal{S}}
\newcommand{\cT}{\mathcal{T}}
\newcommand{\cU}{\mathcal{U}}
\newcommand{\cV}{\mathcal{V}}
\newcommand{\cW}{\mathcal{W}}
\newcommand{\cX}{\mathcal{X}}
\newcommand{\cY}{\mathcal{Y}}
\newcommand{\cZ}{\mathcal{Z}}

\newcommand{\vA}{\mathbf{A}}
\newcommand{\vB}{\mathbf{B}}
\newcommand{\vC}{\mathbf{C}}
\newcommand{\vD}{\mathbf{D}}
\newcommand{\vE}{\mathbf{E}}
\newcommand{\vF}{\mathbf{F}}
\newcommand{\vG}{\mathbf{G}}
\newcommand{\vH}{\mathbf{H}}
\newcommand{\vI}{\mathbf{I}}
\newcommand{\vJ}{\mathbf{J}}
\newcommand{\vK}{\mathbf{K}}
\newcommand{\vL}{\mathbf{L}}
\newcommand{\vM}{\mathbf{M}}
\newcommand{\vN}{\mathbf{N}}
\newcommand{\vO}{\mathbf{O}}
\newcommand{\vP}{\mathbf{P}}
\newcommand{\vQ}{\mathbf{Q}}
\newcommand{\vR}{\mathbf{R}}
\newcommand{\vS}{\mathbf{S}}
\newcommand{\vT}{\mathbf{T}}
\newcommand{\vU}{\mathbf{U}}
\newcommand{\vV}{\mathbf{V}}
\newcommand{\vW}{\mathbf{W}}
\newcommand{\vX}{\mathbf{X}}
\newcommand{\vY}{\mathbf{Y}}
\newcommand{\vZ}{\mathbf{Z}}

\newcommand{\va}{\mathbf{a}}
\newcommand{\vb}{\mathbf{b}}
\newcommand{\vc}{\mathbf{c}}
\newcommand{\vd}{\mathbf{d}}
\newcommand{\ve}{\mathbf{e}}
\newcommand{\vf}{\mathbf{f}}
\newcommand{\vg}{\mathbf{g}}
\newcommand{\vh}{\mathbf{h}}
\newcommand{\vi}{\mathbf{i}}
\newcommand{\vj}{\mathbf{j}}
\newcommand{\vk}{\mathbf{k}}
\newcommand{\vl}{\mathbf{l}}
\newcommand{\vm}{\mathbf{m}}
\newcommand{\vn}{\mathbf{n}}
\newcommand{\vo}{\mathbf{o}}
\newcommand{\vp}{\mathbf{p}}
\newcommand{\vq}{\mathbf{q}}
\newcommand{\vr}{\mathbf{r}}
\newcommand{\Vs}{\mathbf{s}}
\newcommand{\vt}{\mathbf{t}}
\newcommand{\vu}{\mathbf{u}}
\newcommand{\vv}{\mathbf{v}}
\newcommand{\vw}{\mathbf{w}}
\newcommand{\vx}{\mathbf{x}}
\newcommand{\vy}{\mathbf{y}}
\newcommand{\vz}{\mathbf{z}}

\newcommand{\vone}{\mathbf{1}}
\newcommand{\vzero}{\mathbf{0}}

\newcommand{\valpha}{{\boldsymbol{\alpha}}}
\newcommand{\vbeta}{{\boldsymbol{\beta}}}
\newcommand{\vgamma}{{\boldsymbol{\gamma}}}
\newcommand{\vdelta}{{\boldsymbol{\delta}}}
\newcommand{\vepsilon}{{\boldsymbol{\epsilon}}}
\newcommand{\vzeta}{{\boldsymbol{\zeta}}}
\newcommand{\veta}{{\boldsymbol{\eta}}}
\newcommand{\vtheta}{{\boldsymbol{\theta}}}
\newcommand{\viota}{{\boldsymbol{\iota}}}
\newcommand{\vkappa}{{\boldsymbol{\kappa}}}
\newcommand{\vlambda}{{\boldsymbol{\lambda}}}
\newcommand{\vmu}{{\boldsymbol{\mu}}}
\newcommand{\vnu}{{\boldsymbol{\nu}}}
\newcommand{\vxi}{{\boldsymbol{\xi}}}
\newcommand{\vomikron}{{\boldsymbol{\omikron}}}
\newcommand{\vpi}{{\boldsymbol{\pi}}}
\newcommand{\vrho}{{\boldsymbol{\rho}}}
\newcommand{\vsigma}{{\boldsymbol{\sigma}}}
\newcommand{\vtau}{{\boldsymbol{\tau}}}
\newcommand{\vupsilon}{{\boldsymbol{\upsilon}}}
\newcommand{\vphi}{{\boldsymbol{\phi}}}
\newcommand{\vchi}{{\boldsymbol{\chi}}}
\newcommand{\vpsi}{{\boldsymbol{\psi}}}
\newcommand{\vomega}{{\boldsymbol{\omega}}}

\newcommand{\rLambda}{\mathrm{\Lambda}}
\newcommand{\rSigma}{\mathrm{\Sigma}}

\newcommand{\vLambda}{\bm{\rLambda}}
\newcommand{\vSigma}{\bm{\rSigma}}

% big cdot
\makeatletter
\newcommand*\bdot{\mathpalette\bdot@{.7}}
\newcommand*\bdot@[2]{\mathbin{\vcenter{\hbox{\scalebox{#2}{$\m@th#1\bullet$}}}}}
\makeatother

%--------------------------------------------------------------------
% Add a period to the end of an abbreviation unless there's one
% already, then \xspace.
\makeatletter
\DeclareRobustCommand\onedot{\futurelet\@let@token\@onedot}
\def\@onedot{\ifx\@let@token.\else.\null\fi\xspace}

\def\eg{\emph{e.g}\onedot} \def\Eg{\emph{E.g}\onedot}
\def\ie{\emph{i.e}\onedot} \def\Ie{\emph{I.e}\onedot}
\def\cf{\emph{cf}\onedot} \def\Cf{\emph{Cf}\onedot}
\def\etc{\emph{etc}\onedot} \def\vs{\emph{vs}\onedot}
\def\wrt{w.r.t\onedot} \def\dof{d.o.f\onedot} \def\aka{a.k.a\onedot}
\def\etal{\emph{et al}\onedot}
\makeatother

\def\examplecommand{streams\xspace}
\def\modelname{ViTiS\xspace}
\def\up[#1]{\expandafter\MakeUppercase#1}

\newcommand{\relu}{{\operatorname{relu}}}
\newcommand{\tok}{\operatorname{tok}}
\newcommand{\att}{\text{att}}
\newcommand{\ca}{\textsc{ca}}
\newcommand{\sa}{\textsc{sa}}

\newcommand{\best}[1]{{\color{red!60!black}{#1}}}

%%%%%%%%% ABSTRACT
\begin{abstract}
Recent vision-language models are driven by large-scale pretrained models. However, adapting pretrained models on limited data presents challenges such as overfitting, catastrophic forgetting, and the cross-modal gap between vision and language. We introduce a parameter-efficient method to address these challenges, combining multimodal prompt learning and a transformer-based mapping network, while keeping the pretrained models frozen. Our experiments on several video question answering benchmarks demonstrate the superiority of our approach in terms of performance and parameter efficiency on both zero-shot and few-shot settings. Our code is available at \url{https://engindeniz.github.io/vitis}.
\vspace{-2.1em}
\end{abstract}

%%%%%%%%% BODY TEXT
%-------------------------------------------------------------------------
\section{Introduction}
\label{sec:intro}

Recent vision-language models have shown remarkable progress, driven by transformer-based \emph{large-scale pretrained models} ~\cite{dosovitskiy2020image, liu2022vidswin, devlin-etal-2019-bert, roberta, he2021deberta, radford2019language, radford2021learningclip}. These models have been incorporated into video understanding methods, including \emph{video question answering (VideoQA)}, through multimodal fusion on \emph{large-scale multimodal datasets}~\cite{miech2019howto100m, Bain21frozenintime, zellers2021merlot}. However, adapting pretrained models to video-language tasks on limited data is challenging. This is because of the gap between the visual and language modalities and, more importantly, because finetuning the entire model on limited data can lead to overfitting and forgetting previously acquired knowledge.

To address the gap between modalities, transformer-based mapping networks have been employed between frozen vision and language models~\cite{mokady2021clipcap, han2023autoad, alayrac2022flamingo}. These networks map visual features to an appropriate embedding space before they are given as input to the language models. To address overfitting, parameter-efficient adaptation methods have been explored, \eg, \emph{prompt learning}~\cite{li2021prefix, liu2021gpt, liu-etal-2022-p} and \emph{adapter layers}~\cite{houlsby2019adapter} on frozen pretrained models. These approaches preserve the generalization of large-scale models while reducing the number of trainable parameters.

In this work, we investigate the adaptation of large-scale visual-language models to VideoQA under scarcity of training data. Inspired by FrozenBiLM~\cite{yang2022frozenbilm}, we incorporate visual inputs to a frozen language model using lightweight learnable adapter layers. Beyond that, we introduce a novel \emph{visual mapping network} that summarizes the video input while allowing for temporal interaction, inspired by~\cite{mokady2021clipcap, jaegle2021perceiver}. In addition, we introduce \emph{multimodal prompt learning}, which diminishes the number of stored parameters when finetuning in the few-shot setting. We call our model \emph{\textbf{Vi}deoQA with Mul\textbf{ti}-Modal Prompt\textbf{s}} (\modelname).

We pretrain trainable parameters of \modelname, \ie \emph{visual mapping network, adapter layers, visual and text
prompts}, under the \emph{masked language modeling} (MLM) objective on video-text pairs collected from the web, while the vision and language models are kept frozen.
We evaluate \modelname in the zero-shot and few-shot settings.
For the latter, we fine-tune the model on downstream VideoQA tasks, using two approaches: (i) fine-tuning all trainable parameters, which are 8\% of the total model parameters, (ii) fine-tuning only the prompts, which are 0.8\% of all trainable parameters and a mere 0.06\% of the total model parameters.

Our extensive experimental results on multiple open-ended VideoQA datasets demonstrate that \modelname outperforms prior methods while requiring fine-tuning of only a few parameters for each dataset in few-shot settings. In addition, our visual mapping network contributes to better alignment and understanding of multimodal inputs, improving performance in both zero-shot and few-shot settings.

Our contributions can be summarized as follows:
\begin{enumerate}[itemsep=2pt, parsep=0pt, topsep=1pt]
	\item We introduce \emph{multimodal prompt learning} to few-shot VideoQA for the first time, fine-tuning as low as 0.06\% of model parameters on downstream tasks.
	\item We introduce a \emph{visual mapping network} to VideoQA, mapping video input to the text embedding space, while supporting temporal interaction.
	\item We experimentally demonstrate the strong performance of \modelname on multiple VideoQA datasets in both zero-shot and few-shot settings.
\end{enumerate}

\section{Related Work}
\label{sec:related}

\paragraph{Video question answering}

Recent advances in vision-language models benefit from pretrained foundation models, including vision-only~\cite{dosovitskiy2020image, liu2022vidswin} language-only~\cite{devlin-etal-2019-bert, roberta, he2021deberta, radford2019language} and vision-language~\cite{radford2021learningclip}. Recent video understanding methods, including VideoQA, incorporate these models by leveraging large-scale multimodal data~\cite{miech2019howto100m, Bain21frozenintime, zellers2021merlot} with different pretraining objectives, \eg, \emph{masked language modeling}, \emph{masked image modeling}, or \emph{predicting the next word}, to perform single or multiple vision-language tasks~\cite{sun2019videobert, li2020hero, lei2021less, fu2021violet, yang2021justask, zellers2021merlot, li2022align, yang2022frozenbilm, alayrac2022flamingo, zellers2022merlotreserve, cheng2023vindlu, wang2023allinone, li2023lavender, huang2023clover, fu2023empirical}.

Adapting pretrained vision-language models to downstream tasks relies on fully supervised fine-tuning on VideoQA datasets in general~\cite{2016movieqa, xu2017video, jang2017tgif, lei2018tvqa, yu2019activitynet, li2020hero, garcia2020knowit}. Few recent works address the challenge of limited data by focusing on zero-shot~\cite{yang2021justask, yang2022learningjustasktpami, yang2022frozenbilm, alayrac2022flamingo, zellers2022merlotreserve, li2022blip, li2023lavender} and few-shot~\cite{yang2022frozenbilm, alayrac2022flamingo}
open-ended VideoQA tasks. Our work is similar to~\cite{yang2022frozenbilm} in leveraging a frozen video encoder and language model with adapter layers. Beyond that, we introduce a transformer-based visual mapping network between the two models, allowing for temporal interaction. In addition, we incorporate multimodal prompt learning, allowing for efficient fine-tuning in few-shot settings.

%------------------------------------------------------------------------------

\paragraph{Parameter-efficient training}

As the size of large-scale pretrained models grows, adapting them efficiently on limited data without overfitting in an emerging research problem. A common solution is \emph{adapters}, introduced by~\cite{houlsby2019adapter} and employed for vision-language tasks~\cite{eichenberg2022magma, yang2022frozenbilm, sung2022vl}.

Another common solution is \emph{prompting}, referring to inserting tokens to the input to guide pretrained models on downstream tasks. Prompts can be handcrafted (discrete)~\cite{brown2020language} or learned (continuous vectors)~\cite{li2021prefix}. Pretrained language models demonstrate remarkable generalization to zero-shot settings with handcrafted prompts~\cite{brown2020language}. Prompt learning is introduced initially in natural language processing tasks~\cite{li2021prefix, lester-etal-2021-power, liu2021gpt, liu-etal-2022-p, qin-eisner-2021-learning, mahabadi2022promptadapter} and subsequently adopted in vision~\cite{jia2022vpt, bahng2022exploring} and vision-language models. In the latter case, prompts are introduced to text encoders~\cite{zhou2022learningCoOP, zhou2022conditionalCoCoOp}, or both text and vision encoders~\cite{khattak2023maple, wasim2023vita, lee2023multimodal, rasheed2023fine}, called \emph{multimodal}. Learnable prompts can be inserted at the input level~\cite{li2021prefix} and/or deep layers~\cite{liu-etal-2022-p,jia2022vpt}. Few recent works employ prompt learning for video understanding~\cite{ju2022prompting, zhu2022uni, sung2022vl} and multimodal prompt learning for video classification~\cite{wasim2023vita, rasheed2023fine}. We introduce multimodal prompt learning to few-shot VideoQA for the first time.
\section{Method}
\label{sec:method}
%------------------------------------------------------------------------------
\begin{figure*}
\centering
\begin{tabular}{ccc}
	\fig[.45]{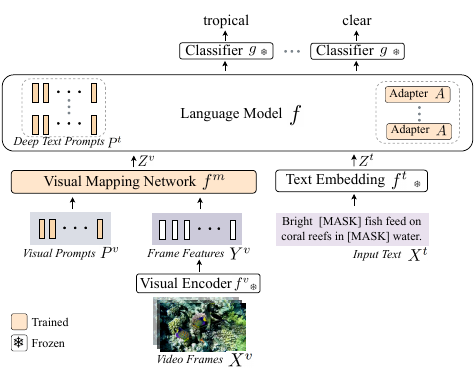} &
	\fig[.25]{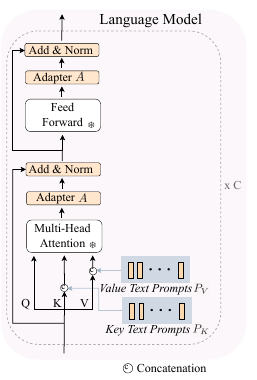} &
	\fig[.25]{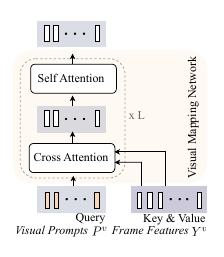} \\
	(a) method overview &
	(b) language model &
	(c) visual mapping network
\end{tabular}
\caption{(a) \modelname consists of a frozen video encoder $f^v$, a visual mapping network $f^m$, a frozen text embedding layer $f^t$, a frozen language model $f$ and a frozen classifier head $g$. Given input video frames $X^v$ and text $X^t$, $f^v$ extracts frame features and $f^m$ maps them to the same space as the text embeddings obtained by $f^t$. Then, $f$ takes the video and text embeddings $Z^v$, $Z^t$ as input and predicts the masked input tokens.
(b) The \emph{language model} incorporates learnable text prompts in the key and value of multi-head-attention and adapter layers after each self-attention and feed-forward layer, before LayerNorm.
(c) Our \emph{visual mapping network} consists of a number of layers, each performing cross-attention between learnable visual prompts and video frame features followed by self-attention.}
\vspace{-1em}
\label{fig:method}
\end{figure*}
%------------------------------------------------------------------------------

The proposed method, \modelname, is illustrated in~\autoref{fig:method}(a), consisting of a frozen video encoder, a visual mapping network, a frozen text embedding layer and a frozen language model that includes learnable text prompts and adapter layers. Given an input video $X^v$, represented as a sequence of frames, and a question $X^q$, represented as a sequence of tokens, the problem is to predict an answer $X^a$ that is another sequence of tokens. The model takes the concatenated sequence $X^t = (X^q, X^a)$ as input text. Parts of $X^t$ may be masked, for example $X^a$ is masked at inference.

%------------------------------------------------------------------------------

\paragraph{Video encoder}

The input video is represented by a sequence of $T$ frames, $X^v = (x_1^v, \dots, x_T^v)$. This sequence is encoded into the \emph{frame features}
\begin{equation}
	Y^v \defn f^v(X^v) = (y_1^v, \dots, y_T^v) \in \real^{D \times T}
\end{equation}
by a frozen pretrained \emph{video encoder} $f^v$, where $D$ is the embedding dimension.

%------------------------------------------------------------------------------

\paragraph{Visual mapping network}

A \emph{visual mapping network} $f^m$ maps the frame features $Y^v$ to the same space as the text embeddings. The mapping is facilitated by a set of $M$ \emph{learnable visual prompts} $P^v \in \real^{D \times M}$, which are given as input along with $Y^v$, to obtain the \emph{video embeddings}
\begin{equation}
	Z^v \defn f^m(P^v, Y^v) \in \real^{D \times M}.
\end{equation}

As shown in \autoref{fig:method}(c), the architecture of $f^m$ is based on Perceiver~\cite{jaegle2021perceiver}, where the latent array corresponds to our learnable visual prompts $P^v$.
It consists of $L$ blocks defined as
\begin{equation}
	Z_\ell \defn \sa_\ell(\ca_\ell(Z_{\ell-1}, Y^v)) \in \real^{D \times M}
\end{equation}
for $\ell = 1, \dots, L$, with input $Z_0 = P^v$. Each block $\ell$ maps the latent vectors $Z_{\ell-1}$ first by cross attention $\ca_\ell$ with the frame features $Y^v$ and then by self-attention $\sa_\ell$ to obtain $Z_\ell$. In cross attention, $Z_{\ell-1}$ serves as query and $Y^v$ as key and value. We thus iteratively extract information from the frame features $Y^v$ into the latent vectors, which are initialized by the visual prompts. The output video embeddings are $Z^v = Z_L \in \real^{D \times M}$. To allow modeling of temporal relations within the video, learnable \emph{temporal position embeddings} are added to $Y^v$ before $f^m$.

%------------------------------------------------------------------------------

\paragraph{Text embedding}

The input text is tokenized into a sequence of $S$ tokens, $X^t = (x_1^t, \dots, x_S^t)$. This sequence is mapped by a frozen \emph{text embedding layer} $f^t$ to the text embedding space,
\begin{equation}
	Z^t \defn f^t(X^t) = (z_1^t, \dots, z_S^t) \in \real^{D \times S}.
\end{equation}
One or more tokens are masked, in which case they are replaced by a learnable mask token.

%------------------------------------------------------------------------------

\paragraph{Language model}

We concatenate video and text embeddings into a single input sequence $(Z^v, Z^t) \in \real^{D \times K}$ of length $K = M + S$. We then feed this sequence to a transformer-based bidirectional language model $f$ to obtain an output sequence
\begin{equation}
	f(Z^v, Z^t) \in \real^{D \times K}
\end{equation}
of the same length. Finally, a classifier head $g$ maps the output sequence to logit vectors over a vocabulary $U$. The logit vectors corresponding to masked tokens are selected to apply the loss function at training or make predictions at inference. Both $f$ and $g$ are pretrained and kept frozen. However, as shown in \autoref{fig:method}(b), $f$ is adapted by means of learnable deep text prompts and adapters, described next.

%-------------------------------------------------------------------------

\paragraph{Text prompts}
\label{sec:text_prompt}

To reduce the number of fine-tuned parameters at downstream tasks, we introduce attention-level text prompts in self-attention blocks at each layer of the language model, referred to as \emph{deep text prompt learning}~\cite{liu-etal-2022-p}. Given a sequence $Z \in \real^{D \times K}$ of token embeddings as input to a self-attention layer of the language model $f$, we prepend two sequences of \emph{learnable text prompts} $P_K, P_V \in \real^{N \times D}$ to the key and value respectively:
\begin{equation}
	Q \defn W_Q Z \quad K \defn [P_K \ \ W_K Z] \quad V \defn [P_V \ \ W_V Z],
\label{eq:attach-text-prompts}
\end{equation}
where $W_Q, W_K, W_V \in \real^{D \times D}$ are the query, key and value projections respectively. The output sequence length does not change since it is determined by the query, where we do not prepend prompts. There is one pair of variables $P_K, P_V$ for each layer of $f$, collectively denoted as $P^t$. These variables are either defined as parameters directly or parametrized by means of projections as discussed next.

\paragraph{Text prompt parametrization}

Instead of defining text prompts as parameters directly, we discuss here an alternative parametrization using projections. We first generate a sequence of input prompts $P^i \in \real ^{D' \times N}$ and then we project it as follows:
\begin{equation}
	P^t \defn W P^i \in \real^{2CD \times N},
\end{equation}
where $W \in \real ^{2CD \times D'}$, $C$ is the number of layers of the language model $f$ and $D$ its embedding dimension. Then, $P^t$ can be reshaped as a $2 \times C \times D \times N$ tensor, representing one pair of sequences $P_K, P_V \in \real^{D \times N}$ for every layer of $f$. After training, the input sequence $P^i$ and projection matrix $W$ are discarded and we only keep $P^t$. This allows us to fine-tune fewer parameters at downstream tasks, which is beneficial when data is limited.

%------------------------------------------------------------------------------

\paragraph{Adapters}

Following~\cite{yang2022frozenbilm}, we add adapter layers to the language model $f$. Given a sequence $Z \in \real^{D \times K}$ of token embeddings, an adapter layer $A$ maps it through a bottleneck dimension $d$ with a residual connection:
\begin{equation}
	A(Z) \defn Z + W_2 h(W_1 Z) \in \real^{D \times K},
\end{equation}
where $W_1 \in \real^{d \times D}$, $W_2 \in \real^{D \times d}$, and $h$ is the $\relu$ activation function. We insert an adapter module after the self-attention layer and the feed-forward layer, preceding LayerNorm in each layer of $f$.

%------------------------------------------------------------------------------

\paragraph{Training and inference}

Our model is trained using the \emph{masked language modeling} (MLM) objective, where one or more tokens of $X^t$ are masked and the corresponding outputs are predicted over a vocabulary $U$. The parameters of the visual encoder $f^v$, text embedding layer $f^t$, language model $f$ and classifier head $g$ are pretrained and kept frozen. Only the newly introduced parameters, that is, visual prompts $P^v$, visual mapping network $f^v$, text prompts $P^t$ and adapter layers, are optimized on video-text pairs. We then fine-tune these parameters or a smaller subset on downstream video question answering tasks, where $X^t = (X^q, X^a)$ consists of a question-answer pair and masking applies to the $X^a$ only. At inference, $X^a$ is masked and the corresponding output yields a prediction.
%-------------------------------------------------------------------------

\section{Experiments}
\label{sec:exp}

\subsection{Datasets}
\label{sec:supp-datasets}

\paragraph{Pretraining}

We use WebVid2M~\cite{Bain21frozenintime} for pretraining, consisting of $2.5$M video-caption pairs scraped from the internet. The domain is open and the captions are manually generated. The average video duration is $18$ seconds and the average caption word count is $12$.

\paragraph{Downstream tasks}

Downstream dataset statistics are given in \autoref{tab:down}. Following~\cite{yang2022frozenbilm}, we use 1\% of the training data for fine-tuning in the few-shot setting.

MSRVTT-QA~\cite{xu2017video} is an extension of MSR-VTT~\cite{xu2016msr}, where question-answer pairs are automatically generated from video descriptions. MSVD-QA~\cite{xu2017video} is based on MSVD~\cite{chen-dolan-2011-collecting} and question-answers pairs are automatically generated as in MSRVTT-QA. ActivityNet-QA~\cite{yu2019activitynet} is derived from ActivityNet~\cite{caba2015activitynet}. The average video duration is $180$s. TGIF-QA~\cite{jang2017tgif} comprises several tasks, including FRAME-QA, where the question can be answered from one of the frames in a GIF. In this work, TGIF-QA refers only to Frame-QA.

\subsection{Implementation details}
\label{sec:supp-impl}

\paragraph{Architecture details}

The \emph{frozen video encoder} is CLIP ViT-L/14~\cite{dosovitskiy2020image,radford2021learningclip}, trained with contrastive loss on $400$M image-text pairs. We uniformly sample $T=10$ frames located at least 1 second apart and each frame is resized to $224 \times 224$ pixels; if the video is shorter than $10$ seconds, we zero-pad up to $T=10$ frames. The encoder then extracts one feature vector per frame of the dimension of $768$, followed by a linear projection to $D=1536$ dimensions.

The \emph{visual mapping network} has $L=2$ layers, each with a cross-attention and a self-attention, having $8$ heads and embedding dimension $D=1536$. We use $M=10$ learnable visual prompt vectors of dimension $D=1536$.

The \emph{text tokenizer} is based on SentencePiece~\cite{kudo-richardson-2018-sentencepiece} with a vocabulary $U$ of size $128$k.

The \emph{frozen language model} is DeBERTa-V2-XLarge~\cite{he2021deberta}, trained using MLM on $160$G text data, following~\cite{yang2022frozenbilm}. The model has $C=24$ layers, $24$ attention heads, and embedding dimension $D=1536$, resulting in $900$M parameters.

%------------------------------------------------------------------------------
\begin{table}[ht!]
\small
\centering
\setlength{\tabcolsep}{3.5pt}
\begin{tabular}{lccccc}
\toprule
\mr{2}{\Th{Dataset}} & \mr{2}{\Th{Videos}} & \mc{4}{\Th{QA Pairs}} \\ \cmidrule{3-6}
& & \Th{Train} & \Th{Val} & \Th{Test} & \Th{Total} \\ \midrule
MSRVTT-QA~\cite{xu2017video}           &  10k & 159k &  12k & 73k &  244k \\
MSVD-QA~\cite{xu2017video}             &   2k &  31k & 6.5k & 13k & 50.5k \\
ActivityNet-QA~\cite{yu2019activitynet} & 5.8k &  32k &  18k &  8k &   58k \\
TGIF-QA~\cite{jang2017tgif}             &  40k &  39k &   -- & 13k &   53k \\
\bottomrule
\end{tabular}
\vspace{-0.5em}
\caption{Downstream dataset statistics.}
\vspace{-1em}
\label{tab:down}
\end{table}
%------------------------------------------------------------------------------

For the \emph{adapter layers}~\cite{houlsby2019adapter}, we set $d = D / 8 = 192$ by following~\cite{yang2022frozenbilm}. For \emph{text prompts}, we use $N=10$ learnable text prompt vectors, $D' = D/8 =192$, and $C=24$.

\paragraph{Downstream input design}

We limit the length of text sequences to $S=256$ tokens for pretraining and zero-shot experiments and $S=128$ tokens for downstream experiments. We adopt the input design of~\cite{yang2022frozenbilm} as follows: "[CLS] Question: $<$Question$>$? Answer: [MASK]. Subtitles: $<$Subtitles$>$ [SEP]". Subtitles are optional and if available, their token sequence $X^s$ is incorporated into the input. In this case, the text input sequence becomes $X^t = (X^q, X^a, X^s)$.

\paragraph{Answer vocabulary}

The answer vocabulary $U$ is constructed by selecting the top 1k most frequent answers from the training set for the zero-shot setting, following~\cite{yang2022frozenbilm,zellers2021merlot}. Another vocabulary is formed by including answers that occur at least twice in the training set for the few-shot setting, as defined in~\cite{yang2022frozenbilm}. Questions with answers outside the vocabulary are excluded from the training process and are assessed as incorrect during evaluation. To report results for the few-shot setting, we choose the vocabulary that yields the best performance on the validation set.

\paragraph{Answer embedding}

The classifier head of the frozen language model includes more tokens than required for downstream training. To address this, by following~\cite{yang2022frozenbilm}, we define a task-specific classification head by keeping the weights of the pretrained head associated with the answer vocabulary. At inference, we provide one mask token at the input, regardless of the ground truth answer length, and we obtain one output logit vector. For multi-token answers, we take the average of the logits corresponding to the ground truth words from the vocabulary.

\paragraph{Evaluation Metrics}

We report top-1 accuracy on public test sets for all downstream tasks, except TGIF-QA where we report on the validation set unless otherwise specified.

%------------------------------------------------------------------------------
\begin{table}
\centering
\small
\setlength{\tabcolsep}{3pt}
\begin{tabular}{lcccccccc}
\toprule
\mr{2}{\#} & \mr{2}{\Th{Ad}} & \mr{2}{\Th{Map}} & \mr{2}{\Th{Pr}} & \Th{Trained} & \Th{Msrvtt} & \Th{Msvd} & \Th{ANet} & \Th{Tgif} \\
  & & & & \Th{Param} & \Th{-QA} & \Th{-QA} & \Th{-QA} & \Th{-QA} \\ \midrule
1 &      & Linear  &     & 1M    & 18.0       & 30.5       & 27.1       & 44.4       \\
2 &      & Linear  & \ch & 15M   & 36.3       & 46.2       & 32.7       & 54.3       \\
3 & \ch  & Linear  &     & 30M   & 35.0       & 45.0       & 32.4       & 53.9       \\
4 & \ch  & Linear  & \ch & 44M   & 36.4       & 47.2       & 32.9       & 54.7       \\ \midrule
5 &      & VPN     &     & 58M   & 24.5       & 37.0       & 26.1       & 50.1       \\
6 &      & VPN     & \ch & 72M   & 36.1       & 47.4       & 34.1       & 55.8       \\
7 & \ch  & VPN     &     & 86M   & 34.7       & 46.0       & 32.4       & 54.4       \\
8 & \ch  & VPN     & \ch & 101M  & \tb{36.5}  & \tb{47.8}  & \tb{37.2}  & \tb{55.9}  \\ \bottomrule
\end{tabular}
\vspace{-0.9em}
\caption{Effect of our proposed components on few-shot top-1 accuracy on the validation set. Pretraining on WebVid2M~\cite{Bain21frozenintime} followed by fine-tuning all trainable parameters on downstream datasets, using 1\% of training data. \Th{Ad}: Adapters; \Th{Map}: mapping network; \Th{Pr}: text prompts; VPN: our visual mapping network. \Th{ANet-QA}: ActivityNet-QA.}
\vspace{-2em}
\label{tab:few-shot-model-ablation}
\end{table}
%------------------------------------------------------------------------------

\paragraph{Training settings}

We use the Adam optimizer~\cite{kingma2014adam} with $\beta=(0.9, 0.95)$ in all experiments. We decay the learning rate using a linear schedule with the warm-up in the first 10\% of the iterations. We use dropout with probability $0.1$ in the language model, adapter layers, text prompts, and visual mapping network. We adopt automatic mixed precision training for all experiments.

We \emph{pretrain} for $10$ epochs on WebVid2M with a batch size of $128$ on $8$ NVIDIA Tesla V100 GPUs, amounting to $20$ hours total training time. The base learning rate is $2 \times 10^{-5}$ and the learning rate for visual and text prompts is separately set to $10^{-3}$.

For \emph{fine-tuning} on each downstream dataset, we train for $20$ epochs with a batch size of $32$ on $4$ NVIDIA Tesla V100 GPUs. The base learning rate is searched over $5$ values in the interval $[10^{-5}, 5 \times 10^{-5}]$, while the learning rate for visual and text prompts is kept at $10^{-3}$. For \emph{prompt-only fine-tuning}, the base learning rate is searched over $3$ values in the interval $[10^{-3}, 3 \times 10^{-3}]$.

%------------------------------------------------------------------------------

\subsection{Ablation}
\label{sec:ablation}

We conduct an ablation study in the few-shot setting.

\paragraph{Model design}

In \autoref{tab:few-shot-model-ablation}, we analyze the effect of different components in the model design.
We observe that changing the baseline from a linear layer to \emph{our visual mapping network} without adapters increases the performance by a large margin in most datasets (row 1$\to$5). By adding \emph{text prompts} to any model design (row 1$\to$2, 3$\to$4, 5$\to$6, 7$\to$8), the performance increases for all datasets. The improvement is vast in the absence of adapters.

The model design that includes a linear mapping network and adapter layers (row 3) corresponds to FrozenBiLM~\cite{yang2022frozenbilm} trained on WebVid2M. While using only our visual mapping network and text prompts (row 6) already works better than FrozenBiLM trained on WebVid2M, we further improve performance by incorporating adapter layers: our full model (row 8) achieves the best performance overall.

\paragraph{Prompt length}

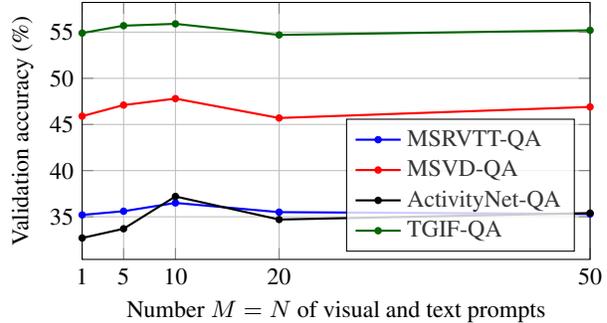
\begin{figure}
\centering
\small
\begin{tikzpicture}
\begin{axis}[
	height=.6\columnwidth,
	xlabel={Number $M = N$ of visual and text prompts},
	ylabel={Validation accuracy (\%)},
	legend pos={south east},
	xmin=1, xmax=50,
	xtick=data,
	xticklabels={1, 5, 10, 20, 50},
]
\pgfplotstableread{
	len msrvtt msvd anet tgif
	1  35.2 45.9 32.7 54.9
	5  35.6 47.1 33.7 55.7
	10  36.5 47.8 37.2 55.9
	20  35.5 45.7 34.7 54.7
	50  35.3 46.9 35.4 55.2
}{\acc}
	\addplot[blue,     mark=*] table[x=len, y=msrvtt] {\acc}; \leg{MSRVTT-QA}
	\addplot[red,      mark=*] table[x=len, y=msvd]   {\acc}; \leg{MSVD-QA}
	\addplot[black,    mark=*] table[x=len, y=anet]   {\acc}; \leg{ActivityNet-QA}
	\addplot[DarkGreen,mark=*] table[x=len, y=tgif]   {\acc}; \leg{TGIF-QA}
\end{axis}
\end{tikzpicture}
\caption{Few-shot top-1 validation accuracy \vs number $M = N$ of \emph{visual and text prompts} for different downstream datasets, using 1\% of training data.}
\label{fig:prompt_len}
\end{figure}

\autoref{fig:prompt_len} shows the effect of the number of prompts on few-shot performance, referring to both visual ($M$) and text ($N$) prompts, \ie, $M = N$. Because the space and time complexity of the model is quadratic in the number of prompts, we limit this number to $50$. We find that accuracy is consistently best on all downstream benchmarks for $M = N = 10$ prompts, which we choose as default.

%------------------------------------------------------------------------------
%------------------------------------------------------------------------------

\paragraph{Number of layers of visual mapping network}

\autoref{tab:vpn-layers} shows the effect of the number $L$ of layers of our visual mapping network on few-shot performance. We only experiment with up to $2$ layers to avoid an excessive number of parameters and complexity of our model. We find that $L = 2$ works best, which we choose as default.

%------------------------------------------------------------------------------
\begin{table}
\centering
\small
\begin{tabular}{ccccc}
\toprule
\Th{VPN} & \Th{Msrvtt} & \Th{Msvd} & \Th{ANet} & \Th{Tgif} \\
\Th{Layers} & \Th{-QA} & \Th{-QA} & \Th{-QA} & \Th{-QA} \\ \midrule
1 & 36.0      & 47.0      & 36.1      & 55.9      \\
2 & \tb{36.5} & \tb{47.8} & \tb{37.2} & \tb{55.9} \\
\bottomrule
\end{tabular}
\caption{Effect of number $L$ of layers of our visual mapping network on few-shot top-1 validation accuracy, using 1\% of training data. \Th{VPN}: Visual Mapping Network. \Th{ANet-QA}: ActivityNet-QA. }
\label{tab:vpn-layers}
\end{table}
% %------------------------------------------------------------------------------
%------------------------------------------------------------------------------
\begin{table}
\centering
\small
\begin{tabular}{ccccc}
\toprule
\mr{2}{\Th{Reparam}} & \Th{Msrvtt} & \Th{Msvd} & \Th{ANet} & \Th{Tgif} \\
& \Th{-QA} & \Th{-QA} & \Th{-QA} & \Th{-QA} \\ \midrule
    & 35.6      & 47.4      & 34.0      & 55.1      \\
\ch & \tb{36.5} & \tb{47.8} & \tb{37.2} & \tb{55.9} \\ \bottomrule
\end{tabular}
\caption{Effect of reparametrization of text prompts on few-shot top-1 validation accuracy, using 1\% of training data. \Th{Reparam}: Reparametrization. \Th{ANet-QA}: ActivityNet-QA.}
\label{tab:few-shot-ablation-texprompt-proj}
\vspace{-2em}
\end{table}
%------------------------------------------------------------------------------

%------------------------------------------------------------------------------
\begin{table*}[ht!]
\centering
\small
\begin{tabular}{clcccc}
\toprule
\mr{2}{\Th{\#}} & \mr{2}{\Th{Input Design}} & \Th{Msrvtt} & \Th{Msvd} & \Th{ANet} & \Th{Tgif} \\
& & \Th{-QA} & \Th{-QA} & \Th{-QA} & \Th{-QA} \\ \midrule
1 & ``[CLS] $<$Question$>$? [MASK]. $<$Subtitles$>$ [SEP]''                                              & 13.2      & 30.2      & 19.8      & 29.8      \\
2 & ``[CLS] \tb{Answer the question:} $<$Question$>$? [MASK]. $<$Subtitles$>$ [SEP]''                    & 7.8       & 22.3      & 14.3      & 35.3      \\
3 & ``[CLS] $<$Question$>$? \tb{Answer:} [MASK]. $<$Subtitles$>$ [SEP]''                                 & 17.7      & 37.2      & \tb{25.8} & 45.1      \\
4 & ``[CLS] \tb{Question:} $<$Question$>$? \tb{Answer:} [MASK].  \tb{Subtitles:} $<$Subtitles$>$ [SEP]'' & \tb{18.0} & \tb{38.2} & 24.9      & \tb{45.5} \\ \bottomrule
\end{tabular}
\caption{Effect of handcrafted prompt placement on \emph{zero-shot} top-1 validation accuracy. \Th{ANet-QA}: ActivityNet-QA.}
\label{tab:zero-shot-ablation-hard-prompt}
\end{table*}
%------------------------------------------------------------------------------

%------------------------------------------------------------------------------
\begin{table*}[ht!]
\centering
\small
\begin{tabular}{clcccc}
\toprule
\mr{2}{\Th{\#}} & \mr{2}{\Th{Input Design}} & \Th{Msrvtt} & \Th{Msvd} & \Th{ANet} & \Th{Tgif} \\
& & \Th{-QA} & \Th{-QA} & \Th{-QA} & \Th{-QA} \\ \midrule
1 & ``[CLS] $<$Question$>$? [MASK]. $<$Subtitles$>$ [SEP]''                                              & 36.3      & 47.0      & 35.8      & 55.8      \\
2 & ``[CLS]\tb{ Answer the question}: $<$Question$>$? [MASK]. $<$Subtitles$>$ [SEP]''                    & 36.3      & 46.8      & 35.1      & 55.8      \\
3 & ``[CLS] $<$Question$>$? \tb{Answer:} [MASK]. $<$Subtitles$>$ [SEP]''                                 & \tb{36.5} & 47.1      & 35.9      & 55.8      \\
4 & ``[CLS] \tb{Question:} $<$Question$>$? \tb{Answer:} [MASK].  \tb{Subtitles:} $<$Subtitles$>$ [SEP]'' & \tb{36.5} & \tb{47.8} & \tb{37.2} & \tb{55.9} \\ \bottomrule
\end{tabular}
\caption{Effect of handcrafted prompt placement on \emph{few-shot} top-1 validation accuracy, using 1\% of training data. \Th{ANet-QA}: ActivityNet-QA.}
\label{tab:few-shot-ablation-hard-prompt}
\end{table*}
%------------------------------------------------------------------------------

\paragraph{Reparametrization of text prompts}

In~\autoref{tab:few-shot-ablation-texprompt-proj}, we investigate the impact of the reparametrization of text prompts, as discussed in \autoref{sec:supp-impl}, on few-shot performance. We find that reparametrization consistently improves performance on all downstream benchmarks. Even though the number of trainable parameters increases from $87$M to $101$M during pretraining and fine-tuning, we do not need to store the additional parameters at inference.

\paragraph{Handcrafted prompts}

We explore the use of handcrafted prompts in the input text. In \autoref{tab:zero-shot-ablation-hard-prompt} and \autoref{tab:few-shot-ablation-hard-prompt}, we consider four different input designs for zero-shot and few-shot settings, respectively: (i) no handcrafted prompts, (ii) placed before the question, (iii) placed just before the [MASK] token (answer), and (iv) placed just before the question, answer and subtitles.

In \emph{zero-shot}, handcrafted prompts are beneficial due to the absence of task-specific learning for downstream tasks. As shown in \autoref{tab:zero-shot-ablation-hard-prompt}, the absence of handcrafted prompts drastically reduces the performance (row 1),  highlighting their necessity. Moreover, the position of the handcrafted prompt has a significant impact on the performance. More specifically, the location of the ``Answer'' prompt affects the results by a large margin (row 2$\to$3), even leading to worse performance than the absence of handcrafted prompts (row 1$\to$2). The presence of an ``Answer'' prompt just before the [MASK] token yields better performance in two input designs (rows 3 \& 4).

Although the impact of using handcrafted text prompts is relatively small in \emph{few-shot} experiments compared to zero-shot experiments, they still improve enhances, particularly on MSRVTT-QA and TGIF-QA datasets, as shown in \autoref{tab:few-shot-ablation-hard-prompt}. Placing handcrafted prompts at the beginning (row 2), as is the case for learnable text prompts, leads to lower performance. The best performance is achieved when handcrafted prompts are placed just before the question, answer, and subtitles (row 4). Therefore, we choose to place handcrafted prompts according to row 4 for both settings.

By contrast, \emph{learable prompts} are all placed at the beginning. We empirically observe that other choices, \eg placing half at the beginning of the input and half just before the [MASK] token, are inferior.

\subsection{Results}
\label{sec:main-results}

\paragraph{Zero-shot}

A comparison with state-of-the-art methods on open-ended zero-shot VideoQA is given in \autoref{tab:zero-shot-sota-extended}. We observe an outstanding performance of our method across different VideoQA benchmarks, despite using significantly less pretraining data compared to other methods. Our performance on ActivityNetQA~\cite{yu2019activitynet} is on par with FrozenBiLM~\cite{yang2022frozenbilm}. Lavender~\cite{li2023lavender} employs a multi-task training approach, transforming different vision-language tasks into MLM. Reserve~\cite{zellers2022merlotreserve} uses GPT-3~\cite{brown2020languagegpt3} to convert questions into masked sentences. Flamingo~\cite{alayrac2022flamingo} uses a frozen auto-regressive language model trained on an extreme-scale dataset. By contrast, our method leverages a lighter frozen language model trained on 2.5M video-text pairs. 

BLIP~\cite{li2022blip} is pretrained on the VQA dataset~\cite{balanced_vqa_v2}, which is not directly comparable as our setting does not involve training on QA pairs. Similarly, Just Ask~\cite{yang2021justask, yang2022learningjustasktpami} uses automatically generated visual question answering datasets. Although these datasets are not annotated by humans, the model is still trained on the specific task. 

%------------------------------------------------------------------------------
\begin{table*}[t!]
\centering
\small
\setlength{\tabcolsep}{4pt}
\begin{tabular}{lcccccccc}
\toprule
\mr{2}{\Th{Method}} & \mr{2}{\Th{Sub}} & \mc{3}{\#\Th{Training}} & \mr{2}{\Th{Msrvtt-QA}} & \mr{2}{\Th{Msvd-QA}} & \mr{2}{\Th{ANet-QA}} & \mr{2}{\Th{Tgif-QA}} \\
& & \Th{Img} & \Th{Vid} & \Th{VQA}  &  &  & &  \\ \midrule

CLIP*~\cite{radford2021learningclip}   &                              & 400M                        & -        &                  & 2.1                                                                    & 7.2                                                                  & 1.2                                                                  & 3.6                                                                  \\
\Th{Reserve~\cite{zellers2022merlotreserve}}  & \ch                                     & -                           & 20M  &                      & 5.8                                                                    & -                                                                    & -                                                                    & -                                                                    \\
\Th{Lavender~\cite{li2023lavender}}    &                                  & 3M                          & 2.5M          &            & 4.5                                                                    & 11.6                                                                 & -                                                                    & 16.7                                                                 \\
Flamingo-3B~\cite{alayrac2022flamingo}    &                               & 2.3B                        & 27M    &                   & 11.0                                                                   & 27.5                                                                 & -                                                           & -                                                                    \\
Flamingo-9B~\cite{alayrac2022flamingo} &                                  & 2.3B                        & 27M     &                  & 13.7                                                                   & 30.2                                                                 & -                                                                    & -                                                                    \\
Flamingo~\cite{alayrac2022flamingo}    &                              & 2.3B                        & 27M          &             & 17.4                                                                   & 35.6                                                                 & -                                                                    & -                                                                    \\
FrozenBiLM~\cite{yang2022frozenbilm}    & \ch                               & -                           & 10M    &                   & 16.7                                                                   & 33.8                                                                 & \textbf{25.9}                                                        & 41.9                                                                 \\ \midrule
\gray{Just Ask~\cite{yang2021justask}}  &             & \gray{69M}  & \gray{-} & \ch & \gray{2.9}                                             & \gray{7.5}                                          & \gray{12.2}                                          & \gray{-}                                             \\
\gray{Just Ask~\cite{yang2022learningjustasktpami}}  &             & \gray{69M}  & \gray{3M} & \ch & \gray{5.6}                                             & \gray{13.5}                                          & \gray{12.3}                                          & \gray{-}                                             \\
\gray{BLIP~\cite{li2022blip}}            &       & \gray{129M} & \gray{-} & \ch & \gray{19.2}                                            & \gray{35.2}                                          & \gray{-}                                             & \gray{-}                                             \\ \midrule
\modelname (Ours)                      &                    & -                           & 2.5M &                      & \textbf{18.2}                                                          & \textbf{36.2}                                                        & 25.0                                                                 & \textbf{45.5} \\
\modelname (Ours)          & \ch                              & -                           & 2.5M  &                    & 18.1                                                          & 36.1                                                      & 25.5                                                                 & \textbf{45.5}                                                        \\ \bottomrule
\end{tabular}%
% }
\caption{\emph{Zero-shot VideoQA} top-1 accuracy on test sets, except TGIF-QA on the validation set. Number of pretraining data: image-text/video-text pairs. \Th{Sub}: subtitle input. \Th{VQA}: visual question answer pairs. \Th{ANet-QA}: ActivityNet-QA. CLIP: CLIP ViT-L/14. Flamingo: Flamingo-80B. We gray out methods trained on VQA pairs, which are not directly comparable. *: CLIP results taken from ~\cite{yang2022frozenbilm}.}
\label{tab:zero-shot-sota-extended}
\end{table*}
%------------------------------------------------------------------------------

%------------------------------------------------------------------------------
\begin{table*}[t!]
\centering
\small
\setlength{\tabcolsep}{4pt}
\newcommand{\std}[1]{{\scriptsize\gray{$\pm$#1}}}
\begin{tabular}{lcccccccc}
\toprule
\mr{2}{\Th{Method}} & \mr{2}{\#\Th{Shot}} & \mc{3}{\#\Th{Pre-Training}} &  \mr{2}{\Th{Msrvtt-QA}} & \mr{2}{\Th{Msvd-QA}} & \mr{2}{\Th{ANet-QA}} & \mr{2}{\Th{Tgif-QA}} \\
& & \Th{Img} & \Th{Vid} & \#\Th{Param}  &  &  &  &  \\ \midrule
Flamingo-3B~\cite{alayrac2022flamingo}  & 32  & 2.3B &  27M & 1.4B & 25.6          & 42.6          & --            & --            \\
Flamingo-9B~\cite{alayrac2022flamingo}  & 32  & 2.3B &  27M & 1.8B & 29.4          & 47.2          & --            & --            \\
Flamingo-80B~\cite{alayrac2022flamingo} & 32  & 2.3B &  27M &  10B & 31.0          & 52.3          & --            & --            \\ \midrule
\modelname (Ours)                                    & 32  &   -- & 2.5M & 101M & 27.0\std{1.0} & 41.9\std{0.8} & 28.7\std{1.3} & 52.2\std{1.2} \\ \bottomrule
\end{tabular}
\caption{\emph{Few-shot VideoQA in-context learning}. Mean and standard deviation of top-1 accuracy on test sets, except TGIF-QA on the validation set, over 10 32-shot tasks drawn at random. Only our model involves parameter updates; we fine-tune 0.75M params. Number of pretraining data: image-text/video-text pairs. \Th{ANet-QA}: ActivityNet-QA.}
\label{tab:few-shot-context-learning}
\end{table*}
%------------------------------------------------------------------------------

\paragraph{Few-shot}

We fine-tune our method on 1\% of the training data by following~\cite{yang2022frozenbilm}, which introduced the few-shot VideoQA task in this form. \autoref{tab:few-shot-sota} compares our method with~\cite{yang2022frozenbilm}. We use two strategies, fine-tuning (i) all trainable parameters and (ii) only prompts. The latter works best, consistently outperforming~\cite{yang2022frozenbilm} while diminishing the number of fine-tuned parameters.

%------------------------------------------------------------------------------
\begin{table}[t!]
\centering
\small
\setlength{\tabcolsep}{2pt}
\resizebox{\columnwidth}{!}{%
\begin{tabular}{lcccccc}
\toprule
\mr{2}{\Th{Method}} & \Th{Trained} & \#\Th{Trained} & \Th{Msrvtt} & \Th{Msvd} & \Th{ANet} & \Th{Tgif} \\
& \Th{Modules} & \Th{Params} & \Th{-QA} & \Th{-QA} & \Th{-QA} & \Th{-QA} \\ \midrule
FrozenBiLM~\cite{yang2022frozenbilm}  & ATP      & 30M    & 36.0       & 46.5       & 33.2       & 55.1       \\
\modelname (Ours)                                  & ATP      & 101M   & 36.5       & 47.6       & 33.1       & 55.7       \\
\modelname (Ours)                                  & Prompts  & 0.75M  & \tb{36.9}  & \tb{47.8}  & \tb{34.2}  & \tb{56.2}  \\ \bottomrule
\end{tabular}%
}
\vspace{-0.9em}
\caption{\emph{Few-shot VideoQA} top-1 accuracy on test sets, except TGIF-QA on the validation set. Number of trained parameters: fine-tuned on the downstream dataset, using 1\% of training data. ATP: All trainable parameters. \Th{ANet-QA}: ActivityNet-QA.}
\label{tab:few-shot-sota}
\end{table}
% %------------------------------------------------------------------------------

\paragraph{Few-shot in-context learning}

An alternative approach for few-shot VideoQA is \emph{in-context learning}~\cite{alayrac2022flamingo}, using few, \eg 32, labeled examples. To compare, we draw $10$ tasks of 32 examples at random from 1\% of training data of each downstream dataset; we fine-tune the prompt vectors, that is, $0.75$M parameters, on each task for 5 epochs and report mean and standard deviation. This can be considered as \emph{test-time prompt tuning}~\cite{shu2022tto} using task-specific annotated data.

\autoref{tab:few-shot-context-learning} shows the results of few-shot in-context learning. Flamingo~\cite{alayrac2022flamingo} uses a frozen auto-regressive language model with trainable cross-attention layers that incorporate vision and language input, trained on an extreme-scale dataset. The Flamingo-3B, Flamingo-9B, and Flamingo-80B  have $1.4$B, $1.8$B, and $10$B learned parameters, respectively, in addition to the frozen language model. By contrast, our method uses a lighter frozen language model and lighter adaptation modules, resulting in only $101$M parameters to learn, and our training data is a relatively small amount of video-text pairs. Despite this, our method outperforms Flamingo-3B~\cite{alayrac2022flamingo} on MSRVTT-QA and is on par with MSVD-QA.
\section{Conclusion}
\label{sec:conclusion}

In this work, we explored the adaptation of large-scale pretrained vision and language models for VideoQA under scarcity of data. We introduced multi-modal prompt learning and a visual mapping network to address challenges in such adaptation. Our method consistently outperforms prior works, while requiring minimal parameter fine-tuning in few-shot VideoQA.

%------------------------------------------------------------------------------

\paragraph{Acknowledgements}
This work was granted access to the HPC resources of IDRIS under the allocation 2022-AD011012263R2 made by GENCI.

%------------------------------------------------------------------------------
\clearpage

{\small
\bibliographystyle{ieee_fullname}
\bibliography{egbib}
}
% %------------------------------------------------------------------------------

\end{document}